\documentclass[letterpaper, 10 pt, conference]{ieeeconf}  
\IEEEoverridecommandlockouts                             
\usepackage{graphicx,scalerel,stackengine}
\usepackage{caption}
\usepackage{subcaption}
\usepackage{multirow}
\usepackage{amssymb}
\usepackage{amsmath}
\usepackage{cite}

\title{\LARGE \bf
A Hierarchical Coding Scheme for Glasses-free 3D Displays Based on Scalable Hybrid Layered Representation of Real-World Light Fields 
}

\author{Joshitha R$^{*}$ and Mansi Sharma$^{*}$
\thanks{$*$ The authors contributed equally to the work.}
\thanks{Joshitha R and Mansi Sharma are with the Department of Electrical Engineering, Indian Institute of Technology Madras, Chennai 600036, India. 
        {\tt\small (email: ee19d401@smail.iitm.ac.in, mansisharmaiitd@gmail.com)}
        }
        }

\begin{document}

\maketitle
\thispagestyle{empty}
\pagestyle{empty}

%%%%%%%%%%%%%%%%%%%%%%%%%%%%%%%%%%%%%%%%%%%%%%%%%%%%%%%%%%%%%%%%%%%%%%%%%%%%%%%%
\begin{abstract}
This paper presents a novel hierarchical coding scheme for light fields based on transmittance patterns of low-rank multiplicative layers and Fourier disparity layers. The proposed scheme learns stacked multiplicative layers from subsets of light field views determined from different scanning orders. The multiplicative layers are optimized using a fast data-driven convolutional neural network (CNN). The essential factor for multiplicative layers representation, which has not been considered in previous compression approaches, is the origin of redundancy, \textit{i.e.}, the low rank structure of light field data. The spatial correlation in layer patterns is exploited with varying low ranks in factorization derived from singular value decomposition on a Krylov subspace. Further, encoding with HEVC efficiently removes intra-view and inter-view correlation in low-rank approximated layers. The initial subset of approximated decoded views from multiplicative representation is used to construct Fourier disparity layer (FDL) representation. The FDL model synthesizes second subset of views which is identified by a pre-defined hierarchical prediction order. The correlations between the prediction residue of synthesized views is further eliminated by encoding the residual signal. The set of views obtained from decoding the residual is employed in order to refine the FDL model and predict the next subset of views with improved accuracy. This hierarchical procedure is repeated until all light field views are encoded. The critical advantage of proposed hybrid layered representation and coding scheme is that it utilizes not just spatial and temporal redundancies, but efficiently exploits the strong intrinsic similarities among neighboring sub-aperture images in both horizontal and vertical directions as specified by different predication orders. Besides, the scheme is flexible to realize a range of multiple bitrates at the decoder within a single integrated system. The compression performance analyzed with real light field shows substantial bitrate savings, maintaining good  reconstruction quality.

\end{abstract}

%%%%%%%%%%%%%%%%%%%%%%%%%%%%%%%%%%%%%%%%%%%%%%%%%%%%%%%%%%%%%%%%%%%%%%%%%%%%%%%%
\section{INTRODUCTION}
\begin{figure}[t!]
    \centering % <-- added
    \begin{subfigure}{0.2\textwidth}
      \includegraphics[width=\linewidth]{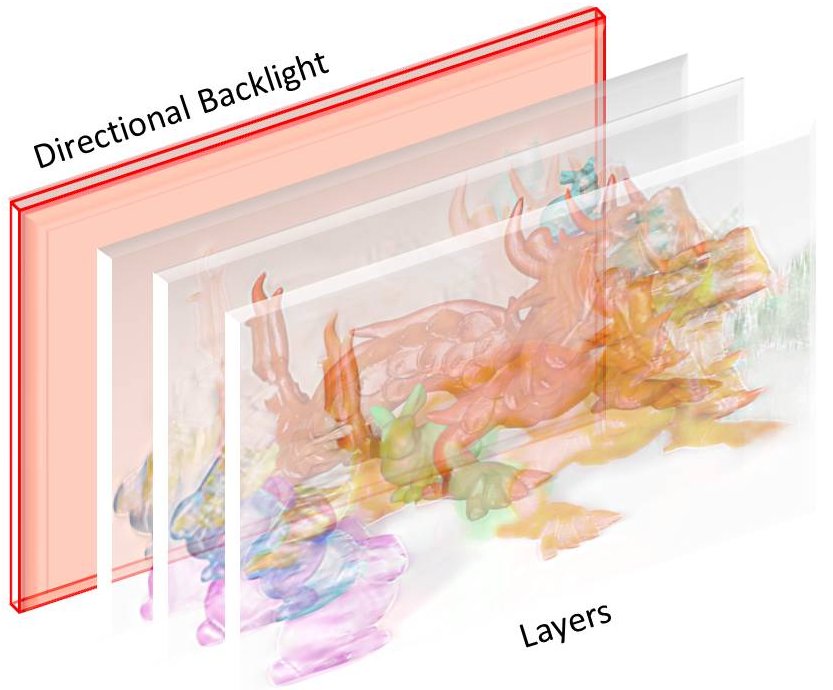}
      \caption{\footnotesize Structure of layered light field display}
      \label{fig:backlight layers}
    \end{subfigure} % <-- added
    \begin{subfigure}{0.21\textwidth}
      \includegraphics[width=\linewidth]{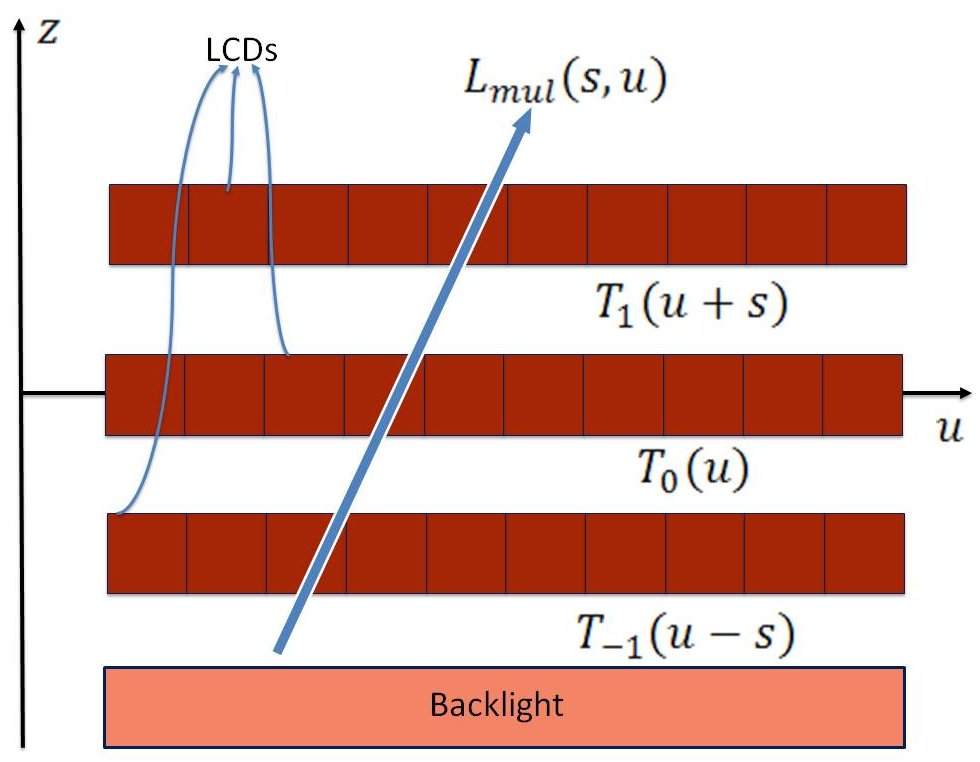}
      \caption{\footnotesize Configuration of multiplicative layers}
      \label{fig:multiplicative layers config}
    \end{subfigure}
\caption{\footnotesize Multiplicative layer patterns for layered light field display.}
\label{fig:layered lfd}
\end{figure}

The market of glasses-free displays is growing rapidly in recent years. Current generation autostereoscopic display techniques are seen as an alternative to stereoscopic 3D as it supports both stereopsis and motion parallax with many views for different viewing directions. However, technology has still not matured enough to simultaneously provide direction-dependent outputs without sacrificing the resolution in reconstructing the dense light fields. Emerging multi-layered light field displays offer solution that provides continuous motion parallax, greater depth-of-field, and wider field-of-view \cite{li2020light, wetzstein2012tensor,sharma2016novel}. A typical structure of a multi-layered display is shown in Fig.~\ref{fig:backlight layers}. It consists of light-attenuating layers stacked in front of a backlight. On each layer, the transmittance of pixels can be controlled independently by carrying out multiplicative operations on the layered patterns. Fig.~\ref{fig:multiplicative layers config} illustrates the light rays which pass through different combinations of pixels in stacked layers depending on the viewing directions. Layered displays accurately reproduce multi-view images or light field simultaneously with high resolution using just few light attenuating layers. Efficient representation and coding of light rays in multi-layered 3D displays is critical for its adaptation on different auto-stereoscopic platforms.

% multiplicative layers with backlight, config

\begin{figure*}[t!]
    \centering 
    \begin{subfigure}{0.56\textwidth}
      \includegraphics[width=\linewidth]{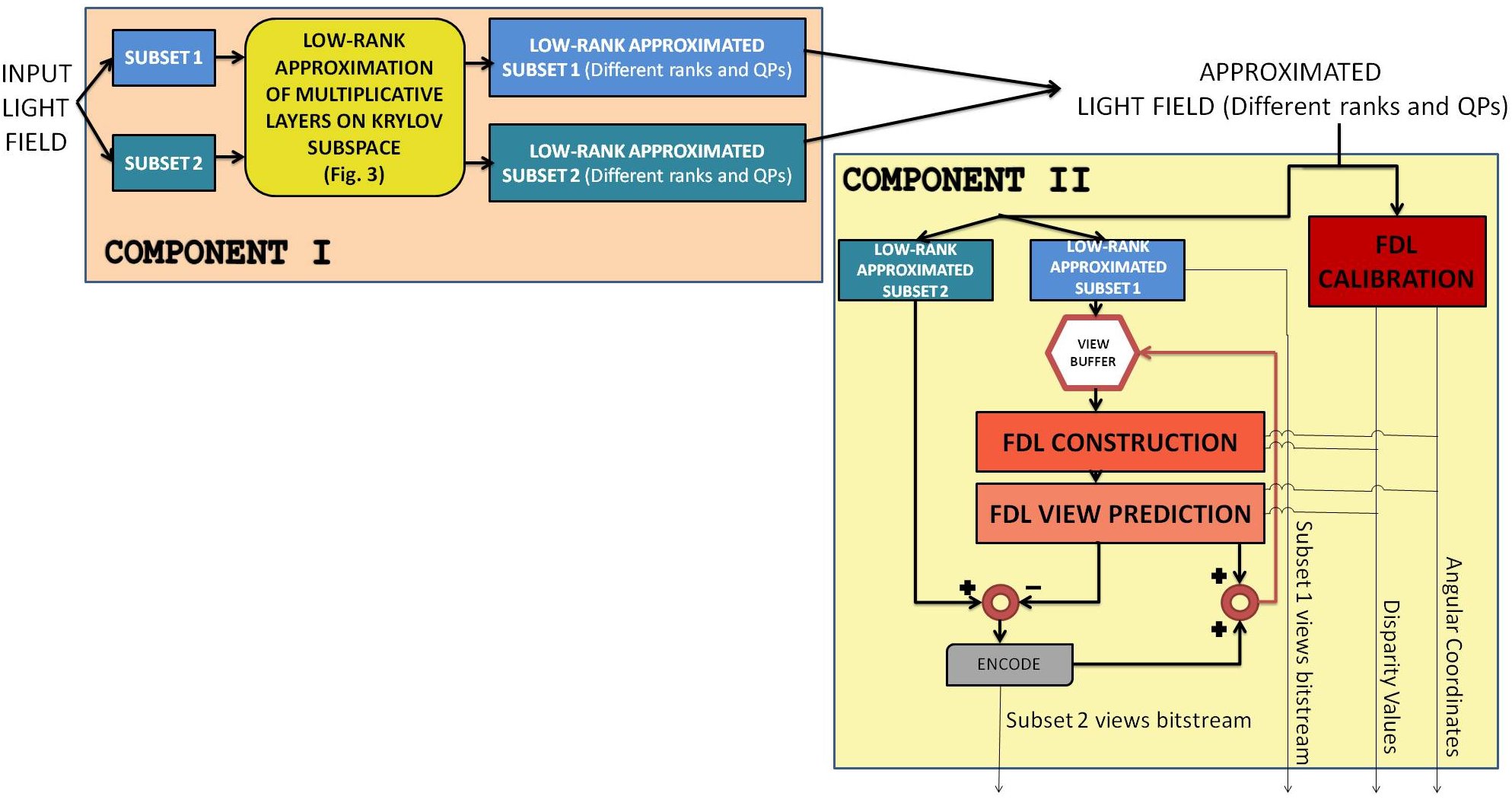}
      \caption{\footnotesize}
      \label{fig:workflow_encoding}
    \end{subfigure}
    \begin{subfigure}{0.35\textwidth}
      \includegraphics[width=\linewidth]{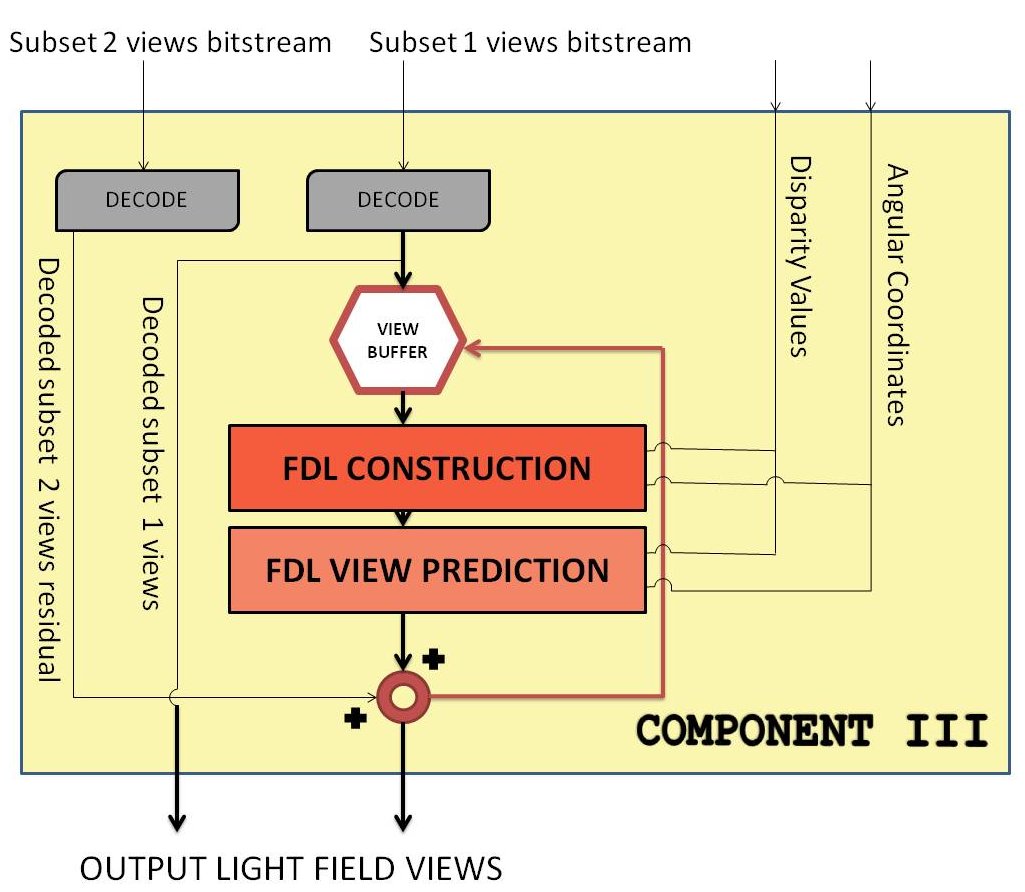}
      \caption{\footnotesize}
      \label{fig:workflow_decoding}
    \end{subfigure}
\caption{\footnotesize The three components of the overall workflow of proposed coding scheme. (a) Overview of the encoding scheme. The low rank approximation of multiplicative layers on Krylov subspace (block in chrome yellow) is expanded in Fig.~\ref{fig:workflow_proposed} (b) Overview of the decoding scheme. }
\label{fig:main workflow}
\end{figure*}
Existing light field coding algorithms are broadly categorized into approaches using the lenslet image \cite{RwlensletRef5_liu2019content,RwlensletRef4_monteiro2017light,RwlensletRef3_li2016compression} or sub-aperture images (SAIs). We can further classify compression methods exploiting geometry information \cite{RwDispRef3_dib2020local,RwDispRef2_jiang2017light,RwDispRef1_zhao2017light,RwEpiRef2_ahmad2020shearlet,RwEpiRef3_chen2020light}, scene content information \cite{RwCbRef1_hu2020adaptive}, view prediction based learning schemes \cite{RwVsRef3_huang2019light,RwVsRef4_heriard2019light, RwDeepRef6_schiopu2019deep,RwDeepRef8_liu2021view} or methods considering light field data as a pseudo video sequence \cite{RwPsvRef1_liu2016pseudo,RwPsvRef3_ahmad2017interpreting,RwPsvRef4_ahmad2019computationally,RwPsvRef5_gu2019high}. The approaches arranging SAIs as pseudo-temporal sequence are constrained to one-dimensional coding structure. Nonlinear correlation in horizontal and vertical directions among adjacent SAIs are not carefully considered in their coding models. The geometry-based or learning-based methods usually demand accurate mathematical model for parameter estimation or suffer from the primary limitation of low quality reconstruction. Their suitability is questionable for compactly encoding light fields with multiple bitrates dealing with layered structure of transmittance patterns in multi-layered displays.

In this paper, we propose a novel hierarchical coding scheme for light fields based on multiplicative layers \cite{maruyama2020comparison} and Fourier disparity layers representation \cite{le2019fourier}. The input light field is divided into two subsets based on predefined Circular and Hierarchical prediction orders (Fig.~\ref{fig:c2h2}). A data-driven convolutional neural network (CNN) is optimized to learn three multiplicative layers for each view subset. The key idea is to reduce dimensionality of stacked multiplicative layers using a randomized block Krylov singular value decomposition (BK-SVD) procedure \cite{musco2015randomized}. Factorization derived from the BK-SVD efficiently exploits the high spatial correlation between multiplicative layers on Krylov subspaces and approximate light fields for varying low ranks. Encoding of these approximated layers using HEVC codec \cite{sullivan2012overview} further eliminates intra-view and inter-view redundancies. In the first stage, the proposed scheme approximates multiplicative layers of a target light field for multiple ranks and the quantization parameters (QPs). The view subsets are reconstructed from the decoded layers.

In the second stage, the entire approximated light field is processed in the Fourier domain following a hierarchical coding procedure. A Fourier Disparity Layer (FDL) calibration estimates disparity values and angular coordinates of each light field view \cite{dib2019light}. These parameters provide additional information for FDL construction and view prediction. These parameters are transmitted to the decoder as metadata. We then split the approximated light field into subsets as identified with scanning and prediction orders. The first set of views is encoded and used for construction of the FDL representation. This FDL representation is used to synthesize the second subset. The remaining correlations between the prediction residue of synthesized views is further eliminated by encoding the residual signal using HEVC with inter coding mode. The set of views obtained from decoding the residual are employed in order to refine the FDL representation and predict the next subset of views with improved accuracy. This hierarchical procedure continues iterating until all light field views are coded. The critical advantages of the proposed hybrid layered representation and coding scheme are:

\begin{itemize} 
\item{It efficiently exploits the strong intrinsic similarities in light field structure by approximating multiplicative layer patterns with varying low ranks. Moreover, it ensures inter-view prediction from adjacent views (both horizontally and vertically) that exhibit higher similarity. The results with different scanning orders demonstrates high compression performance on real world light field data.
}
\item{It is flexible to realize a range of multiple bitrates within a single integrated system trained using few (two) convolutional neural networks. This critical characteristic of the proposed model complements existing light field coding systems or methods which support only specific bitrates during the compression. This would offer an adaptation of layered displays to support a variety of computational or multi-view auto-stereoscopic platforms by optimizing the bandwidth for a given target bitrate.
}

\end{itemize}

\section{PROPOSED SCHEME}

\begin{figure*}[t!]
    \centering
    \includegraphics[scale=0.15]{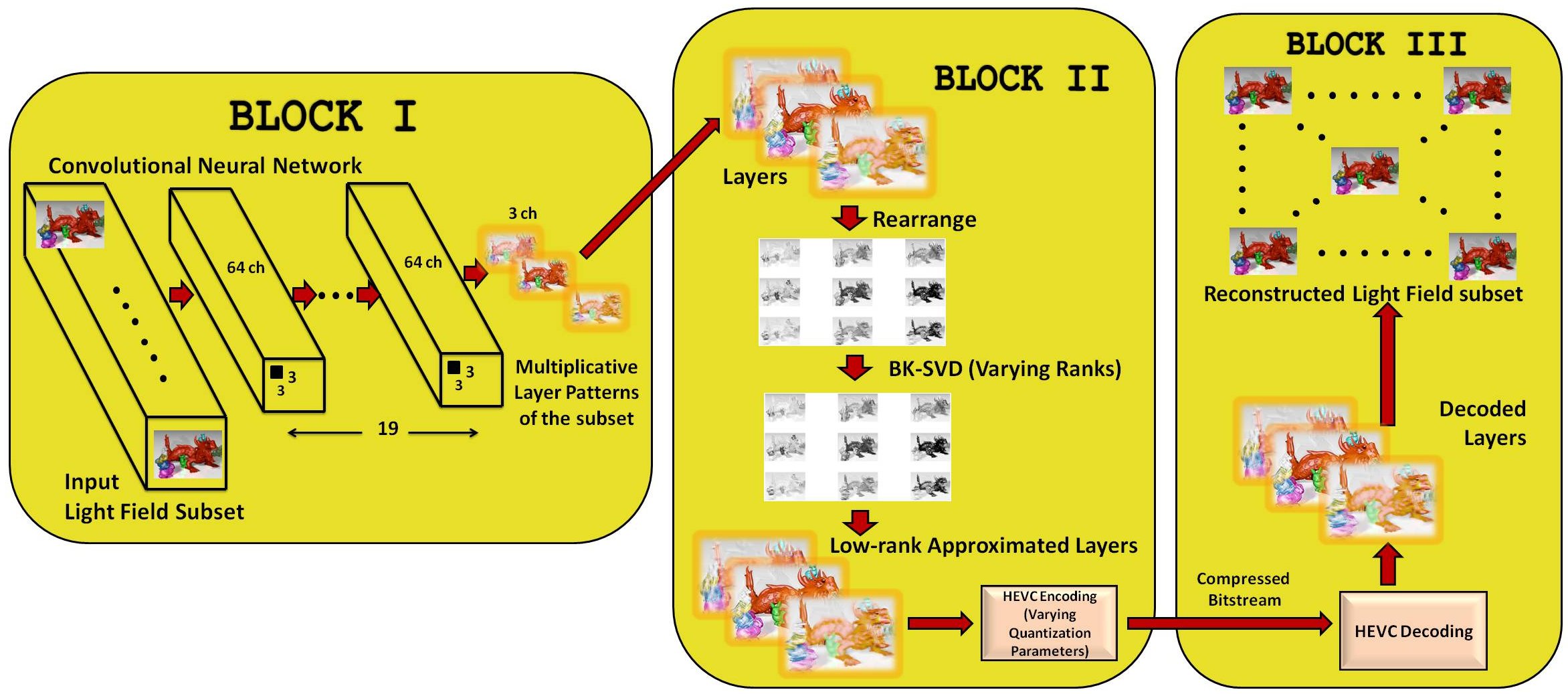}
    \caption{\footnotesize The low rank approximation of multiplicative layers on Krylov subspace consists of three blocks: conversion of light field views into multiplicative layers, low-rank approximation of layers \& HEVC encoding, and the decoding of the approximated layers followed by the reconstruction of the light field.}
    \label{fig:workflow_proposed}
\end{figure*}

The overall workflow of our proposed representation and coding scheme is illustrated in Fig.~\ref{fig:main workflow}. The scheme is divided into three primary components. In \textit{COMPONENT I}, input light field images are divided into subsets, specified by different view prediction orders. The \textit{COMPONENT I} is further expanded into three blocks as depicted in Fig.~\ref{fig:workflow_proposed}. This component performs low-rank approximation of different subsets of the input light field. The \textit{BLOCK I} represents a CNN that converts the light field views of the input subsets into three multiplicative layers \cite{maruyama2020comparison}. In \textit{BLOCK II}, the intrinsic redundancy present in subset views is removed by exploiting the hidden low-rank structure of multiplicative layers on a Krylov subspace for various ranks \cite{musco2015randomized}. HEVC~\cite{sullivan2012overview} encoding of low-rank approximated multiplicative layers further eliminates the inter-frame and intra-frame redundancies in low-rank approximated layers. In \textit{BLOCK III}, the approximated views of the subset are reconstructed from the decoded layers. The entire approximated light field is obtained at the end of \textit{COMPONENT I}.

The approximated subset of decoded views obtained from \textit{COMPONENT I} following multiplicative layer representation is used to construct Fourier disparity layer (FDL)~\cite{le2019fourier} representation in \textit{COMPONENT II}. The processing of approximated light field in \textit{COMPONENT II} of our scheme is shown in Fig.~\ref{fig:workflow_encoding}. There exist strong intrinsic similarities between neighboring sub-aperture images in both horizontal and vertical directions as specified by different predication orders. The main functionality of \textit{COMPONENT II} is to further eliminate inter-view redundancy from low-rank approximated adjacent views as possible determined from different scanning orders. Following a hierarchical procedure, the FDL representation reconstructs the light field iteratively. 

The angular coordinates and disparity values of each view of the low-rank approximated light field (at different ranks and QPs) are determined through FDL calibration and directly transmitted to the decoder (\textit{COMPONENT III}) as metadata~\cite{dib2019light}. The approximated light field is divided into two subsets specified by prediction order. Subset 1 aids in constructing the FDL representation, which is further used to synthesize the second subset of views. The encoding of the residual signal is performed to account for the correlations in the prediction residue of synthesized Subset 2. A more accurate FDL representation is built from previously encoded subsets. Thus, there is an iterative refinement of the FDL representation in \textit{COMPONENT II} until all the approximated light field views are encoded. Our decoding scheme is depicted in the \textit{COMPONENT III} (Fig.~\ref{fig:workflow_decoding}). The angular coordinates and disparity values of the low-rank approximated light field, along with the encoded bitstreams of approximated Subset 1 and Subset 2, are utilized for the final reconstruction.

\subsection{Approximation of light field at different ranks}
\textit{COMPONENT I} illustrates the division of input light field into two subsets and their low-rank approximation in the Block-Krylov subspace. As shown in Fig.~\ref{fig:workflow_proposed}, this step involves three blocks that are described in the following sub-sections.

\subsubsection{View Subsets of Light Field}\label{sec:c2h2}
\begin{figure}[t!]
    \centering 
\begin{subfigure}{0.22\textwidth}
  \includegraphics[width=\linewidth]{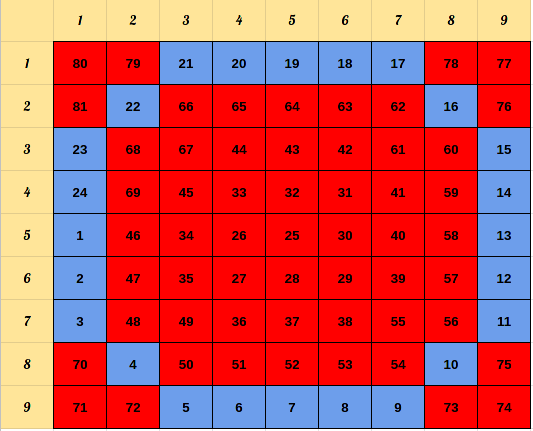}
  \caption{\footnotesize Circular-2}
\end{subfigure}
\begin{subfigure}{0.22\textwidth} 
  \includegraphics[width=\linewidth]{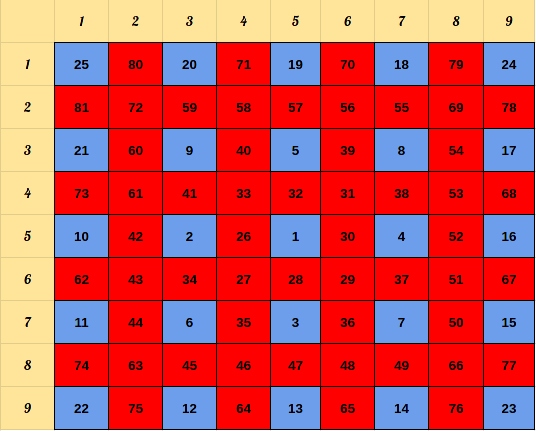}
  \caption{\footnotesize Hierarchical-2}
\end{subfigure} 
\caption{\footnotesize View subsets for the two chosen prediction orders, Circular-2 and Hierarchical-2. Views in blue form the first subset and views in red form the second subset.}
\label{fig:c2h2}
\end{figure}

Dib et al.~\cite{dib2019light} recommend two best scanning orders for synthesizing and coding view subsets, Circular-2 ($C_2$) and Hierarchical-2 ($H_2$). For a 9$\times$9 light field, the $C_2$ pattern contains two subsets. The first subset forms a circular pattern, and the remaining views constitute the second subset. The $H_2$ configuration follows a more simple division of subsets, where each subset comprises of alternate views. The exact coding orders of each subset of the $C_2$ and $H_2$ patterns are shown in Fig.~\ref{fig:c2h2}. In both patterns, the views are ordered beginning with the central view and spiraling out towards the periphery for each subset. Our proposed workflow (Fig.~\ref{fig:main workflow}) begins with partitioning of the input light field into two subsets based on the $C_2$ or $H_2$ pattern, which are then approximated by BK-SVD.

\subsubsection{Light Field Views to Stacked Multiplicative Layers} 

\begin{figure}[t!]
    \centering
    \includegraphics[scale=0.15]{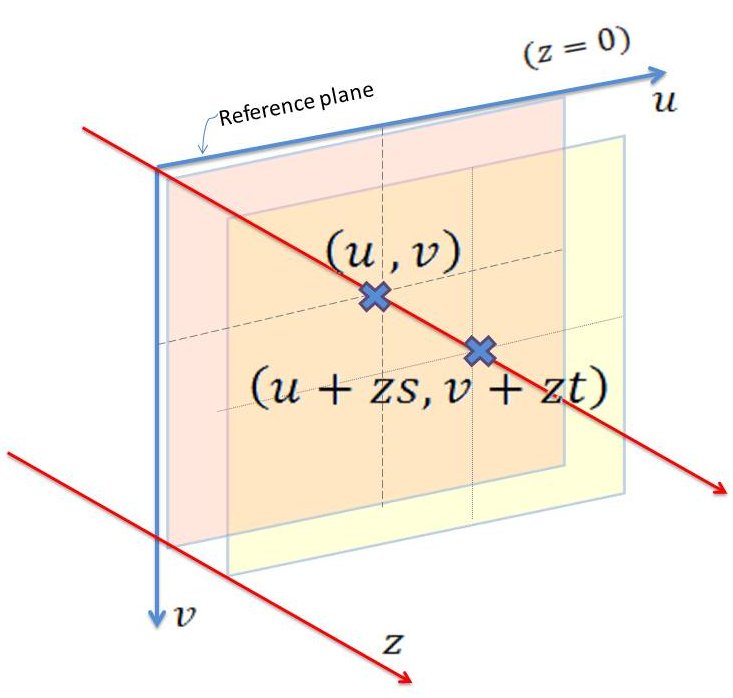}
    \caption{\footnotesize The light ray is parameterized by point of intersection with the $(u,v)$ plane and the $(s,t)$ plane located at a depth $z$.}
    \label{fig:lf planes}
\end{figure}

The four-dimensional light field  $L(u,v,s,t)$ is parameterized by angular coordinates $(u,v)$ and spatial coordinates $(s,t)$ \cite{gortler1996lumigraph,levoy1996light}. If the $(s,t)$ plane is located at depth $z$, the light ray will have coordinates $(u+zs,v+zt)$ (Fig.~\ref{fig:lf planes}). 

Multiplicative layers \cite{maruyama2020comparison} are light attenuating panels that are stacked in evenly spaced intervals in front of a backlight as shown in Fig.~\ref{fig:layered lfd}. The transmittance of a multiplicative layer $M_{z}$ is given by $M_{z}(u,v)$ and a light ray emitted from these layers will have intensity
\begin{equation}
    L_{mul}(u,v,s,t)= \prod_{z \in Z}M_{z}(u+zs,v+zt)
    \label{eq:mult layers}
\end{equation}
where, $z$ is the disparity among directional views and $Z$ denotes a set of $z$. In our work, we considered the two subsets of the light field as $L_{1}(u,v,s,t)$ and $L_{2}(u,v,s,t)$. Three multiplicative layers located at $Z = \{ -1, 0, 1\}$ for each of the two input subsets are obtained. These multiplicative layer patterns are enough to efficiently represent each subset entirely. We performed the optimization of the three layer patterns for each light field by training two CNNs. The final trained network can covert each of the the light field subsets into three optimized output multiplicative layers as depicted in \textit{BLOCK I} of Fig.~\ref{fig:workflow_proposed}.

\subsubsection{Low Rank Representation and Coding of Stacked Multiplicative Layers on Krylov Subspace} \label{sec:bksvd}
The key idea of the proposed scheme is to remove the intrinsic redundancy in light field multiplicative layers by compactly representing them on a Krylov subspace. The randomised Block-Krylov SVD introduced by Cameron Musco and Christopher Musco~\cite{musco2015randomized} can optimally achieve the low-rank approximation of a matrix within $(1 + \epsilon)$ of optimal for spectral norm error. The algorithm quickly converges in $\tilde{O}(\frac{1}{\sqrt{\epsilon}})$ iterations for any matrix. 

In our proposed scheme, we denote each multiplicative layer pattern produced by the CNN as $M_{z} \in \mathbb{R}^{m \times n \times 3}$, where $ z \in \{ -1, 0, 1\}$. The red, green, and blue colour channels of the layer $z$ are denoted as $M^{r}_{z}$, $M^{g}_{z}$, and $M^{b}_{z}$ respectively. We constructed three matrices $B^{ch} \in \mathbb{R}^{3m \times n}$, $ch \in \{ r, g, b \}$, 
as \begin{equation}
B^{ch}= \left( \begin{array}{cc}
    \: \:M^{ch}_{-1} \\ 
    M^{ch}_{0} \\ 
    M^{ch}_{1} \end{array} \right)
\end{equation}
The intrinsic redundancies in multiplicative layers of the light field can be effectively removed by following low-rank BK-SVD approximation in a Krylov subspace for each $B^{ch}$, $ch \in \{ r, g, b \}$. For simplicity, we will denote $B^{ch}$ as $B$ henceforth. Our present formulation of low-rank light field layer approximation seeks to find a subspace that closely captures the variance of $B$'s top singular vectors and avoid the gap dependence in singular values. We target spectral norm low-rank approximation error of $B$, defined as
\begin{equation}
    \left \| B - W W^{T} B  \right \|_{2} \leq \left ( 1 + \epsilon  \right ) \left \|B - B_{k}  \right \|_{2}
    \label{eq:spectral norm error}
\end{equation}
where, $W$ is a rank $k$ matrix with orthonormal columns. In a rank $k$ approximation, only the top $k$ singular vectors of $B$ are considered relevant and the spectral norm guarantee ensures that $W W^{T} B$ recovers $B$ up to a threshold $\epsilon$. We start with a random matrix $\Pi \sim N (0,1)^{d \times k}$ and perform Block Krylov Iteration by working with the Krylov subspace, \begin{equation}
K =[\Pi \hspace{8pt} B_{z}\Pi  \hspace{8pt} B^{2}_{z} \Pi \hspace{8pt} B^{3}_{z}\Pi \cdot \cdot \cdot B^{q}_{z}\Pi ]
\label{eq: Kspace} 
\end{equation}

We constructed $p_{q}(B)\Pi$ for any polynomial $p_{q}(\cdot)$ of degree $q$ by working with Krylov subspace $K$. The approximation of matrix $B$ done by projecting it onto the span of $p_{q}(B)\Pi$, is similar to the best $k$ rank approximation of $B$ lying in the span of the Krylov space $K$. Thus, the nearly optimal low-rank approximation of $B$ can be achieved by finding this best rank $k$ approximation. To ensure that the best spectral norm error low-rank approximation lies in the span of $K$, we orthonormalized the columns of $K$ to obtain $Q \in \mathbb{R}^{c \times qk}$ using the QR decomposition method~\cite{gu1996efficient}. We take SVD of matrix $S = Q^{T} B B^{T} Q$ where, $S \in \mathbb{R}^{qk \times qk}$, for faster computation and accuracy. The rank $k$ approximation of $B$ is matrix $W$, which is obtained as \begin{equation}
    W = Q \bar{U}_{k}
\end{equation}
where, $\bar{U}_{k}$ is set to be the top $k$ singular vectors of $S$. Consequently, the rank $k$ block Krylov approximation of matrices $B^{r}$ , $B^{g}$ and $B^{b}$ are ${W}^{r}$, ${W}^{g}$, and ${W}^{b}$ respectively. The $W^{ch} \in \mathbb{R}^{x \times y}$ for every colour channel $ch$.

To obtain the approximated layers $\widehat{M}_{z} \:,\: z \in \{ -1, 0, 1 \} $, we extract their colour channels from the approximated ${W}^{ch}$ matrices by sectioning out the rows uniformly as 
\begin{subequations}
\begin{equation}
  \widehat{M}^{ch}_{-1} = {W}^{ch}[1 : x \: \:,\: \: 1 : y \: \:]
\end{equation}    
\begin{equation}
  \widehat{M}^{ch}_{0} = {W}^{ch}[x+1 :2x \: \:,\: \: 1 : y \: \:] 
\end{equation}
\begin{equation}
\widehat{M}^{ch}_{1} = {W}^{ch}[2x+1 : 3x \: \:,\: \: 1 : y \: \:] 
\end{equation}
\end{subequations}
The red, green, and blue colour channels are then combined to form each approximated layer $\widehat{M}_{-1}$, $\widehat{M}_{0}$, and $\widehat{M}_{1}$. Thus, factorization derived from the BK-SVD exploits the spatial correlation in multiplicative layers of the subset views for varying low ranks. The three block Krylov approximated layers for each subset are subsequently encoded into a bitstream using HEVC~\cite{sullivan2012overview} for various QPs to further eliminate inter and intra layer redundancies in the low-rank approximated representation. We have shown this low-rank representation and coding of stacked multiplicative layers on Krylov subspace in \textit{BLOCK II} of Fig.~\ref{fig:workflow_proposed}.

%LF Parameterization 

\subsubsection{Decoding and Reconstruction of the Light Field Subsets}
The decoding procedure of compressed layers and the reconstruction of light field subset is illustrated in \textit{BLOCK III} of Fig.~\ref{fig:workflow_proposed}. We decoded three layers from the bitstream and converted them into the RGB format. The decoded layers are denoted as $\grave{M}_{z}$, where $z \in \{ -1, 0, 1 \}$ depicts the depth of the layers. To reconstruct the views of each subset, we considered the integers $s^{*}, t^{*}$ (which depend on the number of views in the subset). The view $I_{(s^{*},t^{*})}$ is obtained by translating the decoded layers to $\bar{\textit{M}}_{z}$ and performing an element-wise product of the colour channels of the translated layers. For a particular view $(s^{*},t^{*})$, the translation of every $z^{th}$ layer $\grave{M}_{z}$, to $\bar{\textit{M}}_{z}$ is carried out as  
\begin{equation}
     \bar{\textit{M}}_{z (s^{*},t^{*})}(u,v) = \grave{M}_{z}(u+zs^{*}, v+zt^{*})
\end{equation}
Thus, the three translated layers for every viewpoint $(s^{*},t^{*})$ are 
\begin{flalign*}
    & \bar{\textit{M}}_{-1(s^{*},t^{*})}(u,v) = \grave{M}_{-1}(u-s^{*}, v-t^{*}) \\
    & \bar{\textit{M}}_{0 (s^{*},t^{*})}(u,v)\:\:\: = \grave{M}_{0}\:(u, v)  \\
    & \bar{\textit{M}}_{1 (s^{*},t^{*})}(u,v)\:\:\: = \grave{M}_{1}(u+s^{*}, v+t^{*})
\end{flalign*}
An element-wise product of each colour channel $ch \in \{r, g, b \}$, of the translated layers gives the corresponding colour channel of the subset view.
\begin{equation}
I^{ch}_{(s^{*},t^{*})}= \bar{\textit{M}}^{ch}_{-1(s^{*},t^{*})} \odot  \bar{\textit{M}}^{ch}_{0(s^{*},t^{*})} \odot \bar{\textit{M}}^{ch}_{1(s^{*},t^{*})}
\end{equation}
The combined red, green, and blue colour channels output the reconstructed light field subset at the viewpoint $(s^{*},t^{*})$ as $I_{(s^{*},t^{*})}$. Subsets 1 and 2 are then merged according to the Circular-2 ($C_2$) or Hierarchical-2 ($H_2$) ordering. 

\textit{COMPONENT I} of the proposed scheme, hence, exploits the spatial correlation in multiplicative layers of the subset views for varying low ranks. Inter and intra layer redundancies in the low-rank approximated representation are removed as well. We further compress the light field by eliminating intrinsic similarities among neighboring views of the $C_2$ or $H_2$ patterns (in both horizontal and vertical directions) using the Fourier Disparity Layers representation.

\subsection{Fourier Disparity Layers (FDL) Representation}
The Fourier Disparity Layers representation \cite{le2019fourier} samples the input BK-SVD approximated light field in the disparity dimension by decomposing it as a discrete sum of layers. The decomposition is carried out in the Fourier domain and FDL representation is constructed by a regularized least square regression performed independently at each spatial frequency. \textit{COMPONENT II} in Fig.~\ref{fig:workflow_encoding} summarises the use of FDL and encoding of low-rank approximated subsets. We highlight the FDL calibration, subset view synthesis and prediction in the following subsections.

\subsubsection{FDL Calibration}
Construction of the FDL requires the angular coordinates $u_{j}$ of the input approximated light field views and the disparity values $d_{k}$ of the layers. These parameters are found by minimizing over all frequency components $f_{s}$ of each view \cite{dib2019light}. The view  $L_{u_{o}}$ of the approximated light field at angular coordinate $u_{o}$ can be defined as $L_{u_{o}}(s)=L(s,u_{o})$ and by using $n$ disparity values $\{d_{k}\}_{k \in [1,n]}$, the Fourier transform of $L_{u_{o}}$ can be written as
\begin{equation}\label{eq:ft}
    \hat{L}_{u_{o}}(f_{s})=\sum_{k} e^{+2i\pi u_{o}d_{k}f_{s}}\hat{L}^{k}(f_{s})
\end{equation}
where, $f_{s}$ is the spatial domain frequency. The Fourier transform of the central light field view obtained by only considering a specific spatial region of disparity $d_{k}$ is given by each $\hat{L}^{k}$. 

By computing the Fourier transforms of all $m$ approximated light field views as $\hat{L}_{u_{j}} (j \in [1,m])$, the FDL representation can be learnt by solving a linear regression problem for each frequency $f_{s}$. The problem is formulated by $\textbf{Ax} = \textbf{b}$ with Tikhonov regularization where $\textbf{A} \in \mathbb{R}^{m \times n}$, $\textbf{x} \in \mathbb{R}^{n \times 1}$ and $\textbf{b} \in \mathbb{R}^{m \times 1}$. Elements of matrix \textbf{A} are $\textbf{A}_{jk} = e^{+2i\pi u_{j}d_{k}f_{s}}$, \textbf{x} contains the Fourier coefficients of disparity layers $\textbf{x}_{k} = \hat{L}^{k}(f_{s})$ and Fourier coefficients of the $j^{th}$ input view, $\textbf{b}_{j} = \hat{L}_{u_{j}}(f_{s})$ are contained in $\textbf{b}$.

\subsubsection{FDL View Synthesis and Prediction}
Any subset view $L_{u_{o}}$ at angular coordinate $u_{o}$ can be synthesized in the Fourier domain with the FDL and their disparities $d_{k}$ using Eq.~\ref{eq:ft}. The $C_2$ and $H_2$ prediction orders divide the low-rank approximated light field into two subsets. The first approximated subset is encoded and aids in constructing the FDL representation. The second approximated subset views are then synthesized using the FDL and the residual signal further refines the representation through iterative encoding of each view. We send the bitstreams of each approximated subsets along with the encoded metadata of parameters to the decoding end, \textit{COMPONENT III} of Fig.~\ref{fig:workflow_decoding}.

\begin{figure}[t!]
    \centering % <-- added
\begin{subfigure}{0.15\textwidth}
  \includegraphics[width=\linewidth]{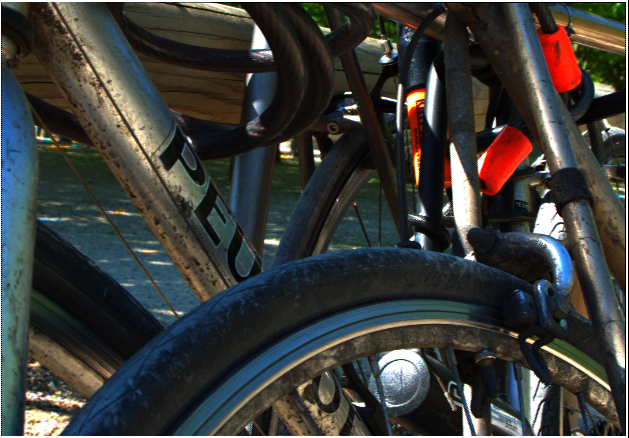}
  \caption{\footnotesize }
\end{subfigure} % <-- added
\begin{subfigure}{0.15\textwidth}
  \includegraphics[width=\linewidth]{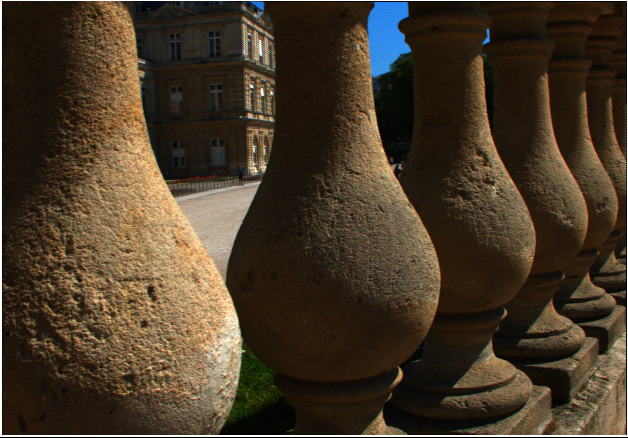}
  \caption{\footnotesize }
\end{subfigure} % <-- added
\begin{subfigure}{0.15\textwidth}
  \includegraphics[width=\linewidth]{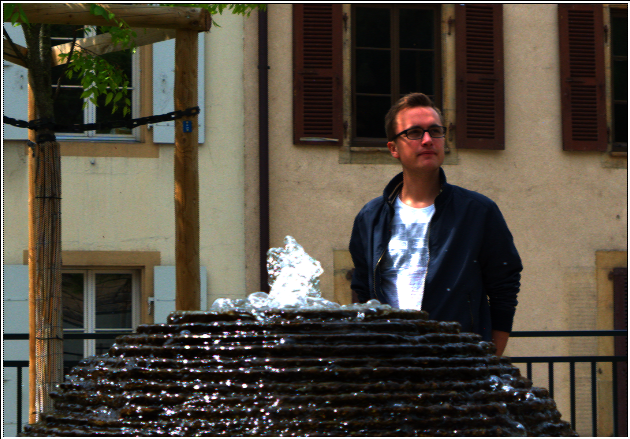}
  \caption{\footnotesize }
\end{subfigure} 
\caption{\footnotesize Central views of the three datasets (a) Bikes, (b) Stone pillars outside, (c) Fountain \& Vincent 2.}
\label{fig:orglfs}
\end{figure}

\section{RESULTS AND ANALYSIS}
The performance of the proposed compression scheme was evaluated on real light fields captured by plenoptic cameras. We experimented with \textit{Bikes}, \textit{Fountain \& Vincent 2}, and \textit{Stone pillars outside} light field datasets from the EPFL Lightfield JPEG Pleno database~\cite{rerabek2016new}. Fig.~\ref{fig:orglfs} shows the central views of the chosen light field images. The raw plenoptic images were extracted into 15$\times$15 sub-aperture views using the Matlab Light field toolbox~\cite{dansereau2013decoding}. The $C_2$ and $H_2$ coding patterns were constructed from the inner 9$\times$9 views for our tests. 

\subsection{Experimental Settings and Implementation Details}\label{sec:exp results}

The proposed scheme was implemented on a single high-end HP OMEN X 15-DG0018TX system with 9th Gen i7-9750H, 16 GB RAM, RTX 2080 8 GB Graphics, and Windows 10 operating system. We optimized the multiplicative layer patterns for each subset using CNN with twenty 2-D convolutional layers stacked in a sequence. The CNN models were trained for 25 epochs at a learning rate of 0.001 and a batch size of 15. The entire network was implemented using the Python-based framework, Chainer (version 7.7.0).

The resultant output multiplicative layers for each subset of the $C_2$ and $H_2$ patterns were obtained from the trained CNN models. We rearranged the colour channels of these multiplicative layers as described in section~\ref{sec:bksvd} and then applied BK-SVD for ranks 4, 8, 16, 28, 44, 52 and 60. The approximated matrices were then arranged back into layers and compressed using HEVC HM 11.0 (\textit{BLOCK II} of \textit{COMPONENT I}). We used quantization parameters 2, 6, 10, 14, 20, 26, and 38 to test both high and low bitrate cases of HEVC. The decoding and reconstruction of BK-SVD approximated subsets for both $C_2$ and $H_2$ patterns was performed.

Low-rank approximated subsets were then utilized to form the FDL representation of the light fields. The number of layers in the FDL method was fixed to $n = 30$. Views in approximated Subset 1 construct the initial FDL representation. Views of approximated Subset 2 are predicted from this FDL representation and the residues iteratively refine the FDL. We used HEVC HM 16.0 to perform the encoding in \textit{COMPONENT II}, for quantization parameters 2, 6, 10, 14, 20, 26 and 38.

\begin{table}[t]
\centering
\caption{\footnotesize The total number of bytes written to file for each subset of the Circular-2 pattern using our proposed coding scheme and Dib et al. \cite{dib2019light}.}
\label{table_bytesC2}
\resizebox{6cm}{!}{
\makebox[\linewidth]{\begin{tabular}{|c|c|c|c|c|c|c|c|}
\hline
\multicolumn{2}{|c|}{} & \multicolumn{2}{|c|}{Bikes} & \multicolumn{2}{|c|}{Stone pillars outside}  & \multicolumn{2}{|c|}{Fountain \& Vincent 2} \\
\hline
QP & Scheme & Subset 1 & Subset 2 & Subset 1 & Subset 2 & Subset 1 & Subset 2\\
\hline
\multirow{8}{*}{2} & Dib et al.	&	6373276	&	9269248	&	6321523	&	7928095	&	6912472	&	8363327	\\	\cline{2-8}
&	Proposed (Rank 4)	&	1634865	&	285016	&	1358448	&	243122	&	1827059	&	335430	\\	\cline{2-8}
&	Proposed (Rank 8)	&	1877550	&	349216	&	1512085	&	283331	&	2115774	&	380847	\\	\cline{2-8}
&	Proposed (Rank 16)	&	2177689	&	417606	&	1847285	&	351680	&	2519102	&	446949	\\	\cline{2-8}
&	Proposed (Rank 28)	&	2438540	&	486569	&	2142502	&	418877	&	2806767	&	494263	\\	\cline{2-8}
&	Proposed (Rank 44)	&	2674089	&	532163	&	2350849	&	487484	&	3008060	&	545947	\\	\cline{2-8}
&	Proposed (Rank 52)	&	2749844	&	561758	&	2401859	&	503039	&	3063768	&	535674	\\	\cline{2-8}
&	Proposed (Rank 60)	&	2805401	&	581882	&	2448047	&	527789	&	3113991	&	535619	\\	\cline{1-8}
\hline
\multirow{8}{*}{6} &	Dib et al.	&	4350810	&	5663692	&	4354754	&	4732850	&	4869817	&	4963030	\\	\cline{2-8}
&	Proposed (Rank 4)	&	602334	&	124074	&	493765	&	104272	&	728339	&	149454	\\	\cline{2-8}
&	Proposed (Rank 8)	&	741293	&	165605	&	571942	&	130267	&	893240	&	178163	\\	\cline{2-8}
&	Proposed (Rank 16)	&	916846	&	211107	&	761895	&	173837	&	1138839	&	220365	\\	\cline{2-8}
&	Proposed (Rank 28)	&	1086119	&	247005	&	925727	&	218830	&	1349095	&	252514	\\	\cline{2-8}
&	Proposed (Rank 44)	&	1252891	&	282873	&	1051703	&	256343	&	1499318	&	268165	\\	\cline{2-8}
&	Proposed (Rank 52)	&	1311143	&	290011	&	1084796	&	268153	&	1547468	&	279455	\\	\cline{2-8}
&	Proposed (Rank 60)	&	1354933	&	310371	&	1118995	&	281491	&	1583718	&	284981	\\	\cline{1-8}
\hline
\multirow{8}{*}{10} &	Dib et al.	&	4350810	&	5663692	&	4354754	&	4732850	&	4869817	&	4963030	\\	\cline{2-8}
&	Proposed (Rank 4)	&	602334	&	124074	&	493765	&	104272	&	728339	&	149454	\\	\cline{2-8}
&	Proposed (Rank 8)	&	741293	&	165605	&	571942	&	130267	&	893240	&	178163	\\	\cline{2-8}
&	Proposed (Rank 16)	&	916846	&	211107	&	761895	&	173837	&	1138839	&	220365	\\	\cline{2-8}
&	Proposed (Rank 28)	&	1086119	&	247005	&	925727	&	218830	&	1349095	&	252514	\\	\cline{2-8}
&	Proposed (Rank 44)	&	1252891	&	282873	&	1051703	&	256343	&	1499318	&	268165	\\	\cline{2-8}
&	Proposed (Rank 52)	&	1311143	&	290011	&	1084796	&	268153	&	1547468	&	279455	\\	\cline{2-8}
&	Proposed (Rank 60)	&	1354933	&	310371	&	1118995	&	281491	&	1583718	&	284981	\\	\cline{1-8}
\hline
\multirow{8}{*}{14} &	Dib et al.	&	1635624	&	1555027	&	1697516	&	1227782	&	1879832	&	1231959	\\	\cline{2-8}
&	Proposed (Rank 4)	&	85418	&	47842	&	60119	&	36583	&	124338	&	51127	\\	\cline{2-8}
&	Proposed (Rank 8)	&	129326	&	64433	&	81540	&	45979	&	177036	&	66168	\\	\cline{2-8}
&	Proposed (Rank 16)	&	197901	&	80959	&	136933	&	64691	&	279887	&	79125	\\	\cline{2-8}
&	Proposed (Rank 28)	&	271565	&	98486	&	200957	&	79562	&	376807	&	94429	\\	\cline{2-8}
&	Proposed (Rank 44)	&	347606	&	112747	&	255491	&	93448	&	449040	&	103222	\\	\cline{2-8}
&	Proposed (Rank 52)	&	372263	&	119795	&	267999	&	97639	&	466826	&	103998	\\	\cline{2-8}
&	Proposed (Rank 60)	&	389902	&	127835	&	280325	&	102454	&	483184	&	107398	\\	\cline{1-8}
\hline
\multirow{8}{*}{20} &	Dib et al.	&	1635624	&	1555027	&	695581	&	292728	&	713369	&	335902	\\	\cline{2-8}
&	Proposed (Rank 4)	&	85418	&	47842	&	25871	&	19602	&	54329	&	25826	\\	\cline{2-8}
&	Proposed (Rank 8)	&	129326	&	64433	&	34316	&	23444	&	76408	&	33557	\\	\cline{2-8}
&	Proposed (Rank 16)	&	197901	&	80959	&	55115	&	31584	&	116924	&	40566	\\	\cline{2-8}
&	Proposed (Rank 28)	&	271565	&	98486	&	77665	&	38192	&	160911	&	47636	\\	\cline{2-8}
&	Proposed (Rank 44)	&	347606	&	112747	&	99503	&	44221	&	187776	&	51037	\\	\cline{2-8}
&	Proposed (Rank 52)	&	372263	&	119795	&	104642	&	46549	&	195728	&	52069	\\	\cline{2-8}
&	Proposed (Rank 60)	&	389902	&	127835	&	109416	&	48116	&	203384	&	51973	\\	\cline{1-8}
\hline
\multirow{8}{*}{26} &	Dib et al.	&	191049	&	125748	&	180485	&	36164	&	213553	&	66399	\\	\cline{2-8}
&	Proposed (Rank 4)	&	18109	&	15295	&	12194	&	10550	&	26763	&	14884	\\	\cline{2-8}
&	Proposed (Rank 8)	&	26471	&	20342	&	16372	&	13010	&	35126	&	18824	\\	\cline{2-8}
&	Proposed (Rank 16)	&	36035	&	24566	&	22220	&	15525	&	46554	&	19651	\\	\cline{2-8}
&	Proposed (Rank 28)	&	46328	&	28090	&	28818	&	17922	&	60959	&	23291	\\	\cline{2-8}
&	Proposed (Rank 44)	&	55291	&	29746	&	35061	&	20507	&	70030	&	24332	\\	\cline{2-8}
&	Proposed (Rank 52)	&	58525	&	30969	&	37056	&	20533	&	73670	&	24606	\\	\cline{2-8}
&	Proposed (Rank 60)	&	61024	&	33335	&	39291	&	21200	&	75358	&	25193	\\	\cline{1-8}
\hline
\multirow{8}{*}{38} &	Dib et al.	&	23065	&	7880	&	12613	&	1880	&	22582	&	2155	\\	\cline{2-8}
&	Proposed (Rank 4)	&	2737	&	3283	&	2289	&	3043	&	5098	&	3887	\\	\cline{2-8}
&	Proposed (Rank 8)	&	4208	&	4244	&	2719	&	3314	&	6606	&	4240	\\	\cline{2-8}
&	Proposed (Rank 16)	&	5752	&	5095	&	3197	&	3572	&	7206	&	4708	\\	\cline{2-8}
&	Proposed (Rank 28)	&	6864	&	5757	&	3888	&	3914	&	8261	&	4785	\\	\cline{2-8}
&	Proposed (Rank 44)	&	7852	&	6246	&	4442	&	3969	&	9137	&	5104	\\	\cline{2-8}
&	Proposed (Rank 52)	&	7946	&	6244	&	4437	&	4024	&	9294	&	5126	\\	\cline{2-8}
&	Proposed (Rank 60)	&	8559	&	6592	&	4553	&	4038	&	9630	&	5197	\\	\cline{1-8}
\hline
\end{tabular} }
}
\end{table}

%table2-bytesh2
\begin{table}[t]
\centering
\caption{\footnotesize Total number of bytes for each subset of Hierarchical-2 pattern using our proposed coding scheme and Dib et al. \cite{dib2019light}. }
\label{table_bytesH2}
\resizebox{6cm}{!}{
\makebox[\linewidth]{\begin{tabular}{|c|c|c|c|c|c|c|c|}
\hline
\multicolumn{2}{|c|}{} & \multicolumn{2}{|c|}{Bikes} & \multicolumn{2}{|c|}{Stone pillars outside}  & \multicolumn{2}{|c|}{Fountain \& Vincent 2} \\
\hline
QP & Scheme & Subset 1 & Subset 2 & Subset 1 & Subset 2 & Subset 1 & Subset 2\\
\hline
\multirow{8}{*}{2} &	Dib et al.	&	8149470	&	9362963	&	8187441	&	7746889	&	8467600	&	8311549	\\	\cline{2-8}
&	Proposed (Rank 4)	&	1752364	&	327571	&	1544489	&	277441	&	2000768	&	399280	\\	\cline{2-8}
&	Proposed (Rank 8)	&	1923491	&	425176	&	1668430	&	336414	&	2197190	&	475908	\\	\cline{2-8}
&	Proposed (Rank 16)	&	2132305	&	534173	&	1904622	&	452160	&	2413070	&	576114	\\	\cline{2-8}
&	Proposed (Rank 28)	&	2307124	&	626191	&	2137972	&	550384	&	2589097	&	663419	\\	\cline{2-8}
&	Proposed (Rank 44)	&	2466765	&	728164	&	2348147	&	638116	&	2701529	&	709257	\\	\cline{2-8}
&	Proposed (Rank 52)	&	2531862	&	732311	&	2415703	&	670368	&	2728626	&	732779	\\	\cline{2-8}
&	Proposed (Rank 60)	&	2572006	&	759606	&	2470431	&	700779	&	2741756	&	740021	\\	\cline{1-8}
\hline
\multirow{8}{*}{6} &	Dib et al.	&	5968201	&	5584732	&	5943824	&	4566311	&	6317422	&	4734277	\\	\cline{2-8}
&	Proposed (Rank 4)	&	696307	&	141510	&	600337	&	110556	&	870560	&	169079	\\	\cline{2-8}
&	Proposed (Rank 8)	&	796855	&	191515	&	665513	&	143058	&	991937	&	208585	\\	\cline{2-8}
&	Proposed (Rank 16)	&	928516	&	259814	&	809014	&	209917	&	1119092	&	266975	\\	\cline{2-8}
&	Proposed (Rank 28)	&	1044367	&	309978	&	953162	&	267569	&	1237456	&	314759	\\	\cline{2-8}
&	Proposed (Rank 44)	&	1151050	&	364678	&	1079187	&	314645	&	1320242	&	341091	\\	\cline{2-8}
&	Proposed (Rank 52)	&	1195073	&	376134	&	1122665	&	333719	&	1345621	&	355264	\\	\cline{2-8}
&	Proposed (Rank 60)	&	1224508	&	388297	&	1153867	&	350318	&	1356529	&	359455	\\	\cline{1-8}
\hline
\multirow{8}{*}{10} &	Dib et al.	&	3906437	&	2991907	&	3972408	&	2394424	&	4256777	&	2365804	\\	\cline{2-8}
&	Proposed (Rank 4)	&	178206	&	74529	&	139283	&	54783	&	278393	&	83747	\\	\cline{2-8}
&	Proposed (Rank 8)	&	234038	&	103767	&	176282	&	74433	&	350782	&	106751	\\	\cline{2-8}
&	Proposed (Rank 16)	&	313521	&	143968	&	251942	&	112280	&	442029	&	138069	\\	\cline{2-8}
&	Proposed (Rank 28)	&	389826	&	175682	&	336035	&	146248	&	523367	&	166752	\\	\cline{2-8}
&	Proposed (Rank 44)	&	465603	&	213126	&	423251	&	172780	&	582988	&	184091	\\	\cline{2-8}
&	Proposed (Rank 52)	&	493156	&	213184	&	449828	&	182623	&	600907	&	189758	\\	\cline{2-8}
&	Proposed (Rank 60)	&	515966	&	221252	&	476680	&	192152	&	606755	&	192677	\\	\cline{1-8}
\hline
\multirow{8}{*}{14} &	Dib et al.	&	2265550	&	1479998	&	2432694	&	1068158	&	2505792	&	1058776	\\	\cline{2-8}
&	Proposed (Rank 4)	&	80756	&	43910	&	56613	&	32851	&	125237	&	48892	\\	\cline{2-8}
&	Proposed (Rank 8)	&	108524	&	61888	&	74445	&	41825	&	162045	&	61496	\\	\cline{2-8}
&	Proposed (Rank 16)	&	150025	&	85871	&	111220	&	63932	&	212590	&	78435	\\	\cline{2-8}
&	Proposed (Rank 28)	&	191702	&	105961	&	151212	&	82963	&	264760	&	93719	\\	\cline{2-8}
&	Proposed (Rank 44)	&	228876	&	121783	&	194620	&	96725	&	297942	&	104203	\\	\cline{2-8}
&	Proposed (Rank 52)	&	243903	&	126699	&	209660	&	101918	&	309631	&	106855	\\	\cline{2-8}
&	Proposed (Rank 60)	&	254844	&	129017	&	222970	&	106454	&	314274	&	109546	\\	\cline{1-8}
\hline
\multirow{8}{*}{20} &	Dib et al.	&	842135	&	459805	&	946011	&	221956	&	926805	&	286174	\\	\cline{2-8}
&	Proposed (Rank 4)	&	31602	&	20923	&	21663	&	15803	&	50142	&	23547	\\	\cline{2-8}
&	Proposed (Rank 8)	&	43511	&	29373	&	28169	&	19788	&	63745	&	29074	\\	\cline{2-8}
&	Proposed (Rank 16)	&	60423	&	40866	&	40732	&	28521	&	83392	&	35879	\\	\cline{2-8}
&	Proposed (Rank 28)	&	76647	&	49200	&	53703	&	35594	&	103900	&	42129	\\	\cline{2-8}
&	Proposed (Rank 44)	&	89653	&	56477	&	66649	&	41210	&	118608	&	45865	\\	\cline{2-8}
&	Proposed (Rank 52)	&	94969	&	59779	&	71693	&	42630	&	121445	&	47138	\\	\cline{2-8}
&	Proposed (Rank 60)	&	98186	&	61299	&	75923	&	44694	&	124309	&	47441	\\	\cline{1-8}
\hline
\multirow{8}{*}{26} &	Dib et al.	&	242763	&	111266	&	213782	&	28940	&	272168	&	50062	\\	\cline{2-8}
&	Proposed (Rank 4)	&	15705	&	11349	&	9646	&	7203	&	20645	&	11951	\\	\cline{2-8}
&	Proposed (Rank 8)	&	20006	&	14619	&	12592	&	9004	&	27722	&	14167	\\	\cline{2-8}
&	Proposed (Rank 16)	&	26014	&	19172	&	15947	&	12784	&	32178	&	15899	\\	\cline{2-8}
&	Proposed (Rank 28)	&	30615	&	22252	&	19394	&	14768	&	39019	&	17819	\\	\cline{2-8}
&	Proposed (Rank 44)	&	34692	&	25694	&	23190	&	16965	&	42684	&	19317	\\	\cline{2-8}
&	Proposed (Rank 52)	&	36820	&	26082	&	25005	&	17802	&	44419	&	19885	\\	\cline{2-8}
&	Proposed (Rank 60)	&	37464	&	26717	&	26359	&	18185	&	45207	&	19866	\\	\cline{1-8}
\hline
\multirow{8}{*}{38} &	Dib et al.	&	26295	&	6202	&	13132	&	2001	&	22971	&	1975	\\	\cline{2-8}
&	Proposed (Rank 4)	&	2485	&	2434	&	2065	&	2188	&	4019	&	3065	\\	\cline{2-8}
&	Proposed (Rank 8)	&	3475	&	3205	&	2465	&	2484	&	4976	&	3523	\\	\cline{2-8}
&	Proposed (Rank 16)	&	4143	&	3742	&	2736	&	2707	&	5456	&	3699	\\	\cline{2-8}
&	Proposed (Rank 28)	&	4872	&	4364	&	3002	&	2868	&	6114	&	4137	\\	\cline{2-8}
&	Proposed (Rank 44)	&	5280	&	4572	&	3121	&	2984	&	6327	&	4141	\\	\cline{2-8}
&	Proposed (Rank 52)	&	5604	&	4810	&	3174	&	2980	&	6384	&	4192	\\	\cline{2-8}
&	Proposed (Rank 60)	&	5704	&	4866	&	3205	&	3033	&	6467	&	4291	\\	\cline{1-8}
\hline
\end{tabular} }
}
\end{table}

\subsection{Results and Comparative Analysis }

\begin{figure}[t!]
    \centering % <-- added
\begin{subfigure}{0.21\textwidth}
  \includegraphics[width=\linewidth]{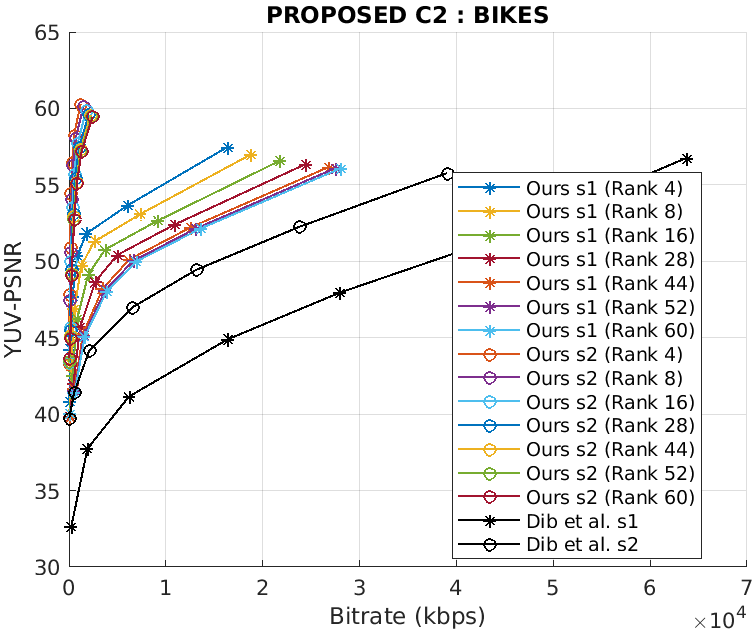}
\end{subfigure} % <-- added
\begin{subfigure}{0.21\textwidth}
  \includegraphics[width=\linewidth]{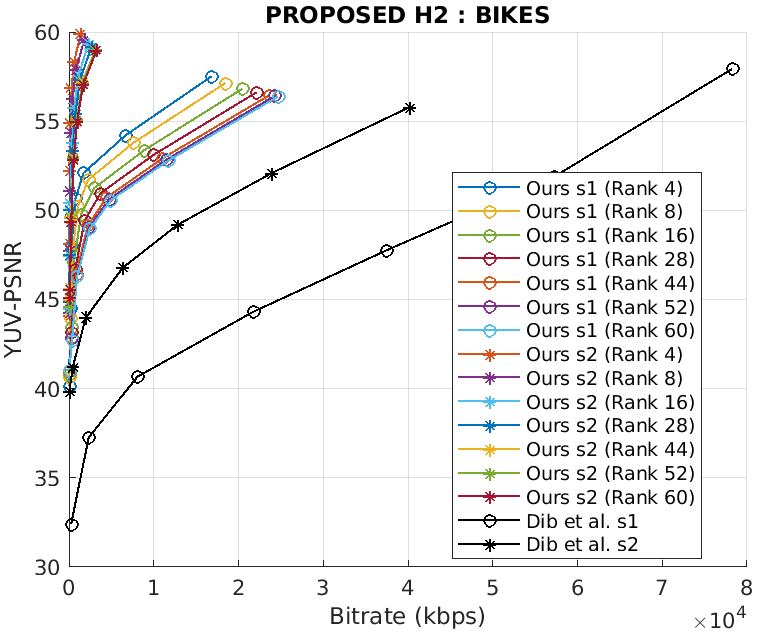}
\end{subfigure} % <-- added

\begin{subfigure}{0.21\textwidth}
  \includegraphics[width=\linewidth]{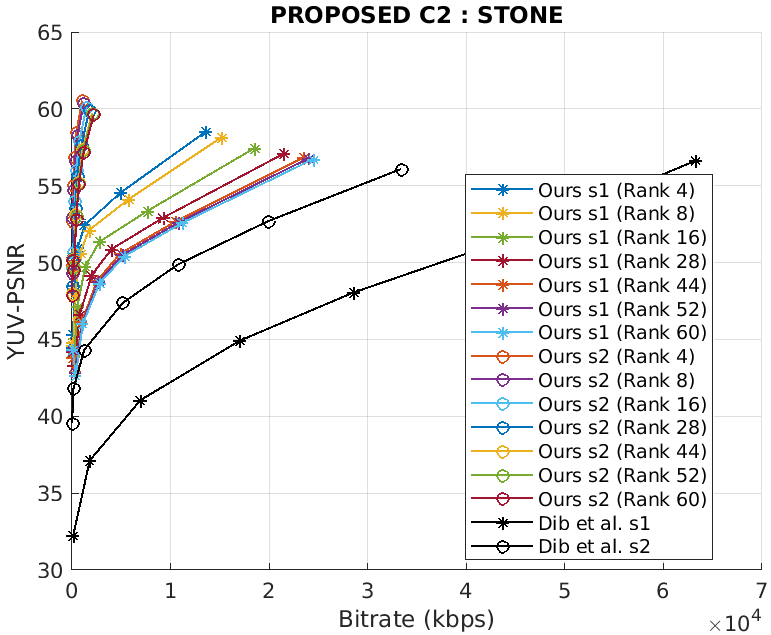}
\end{subfigure} % <-- added
\begin{subfigure}{0.21\textwidth}
  \includegraphics[width=\linewidth]{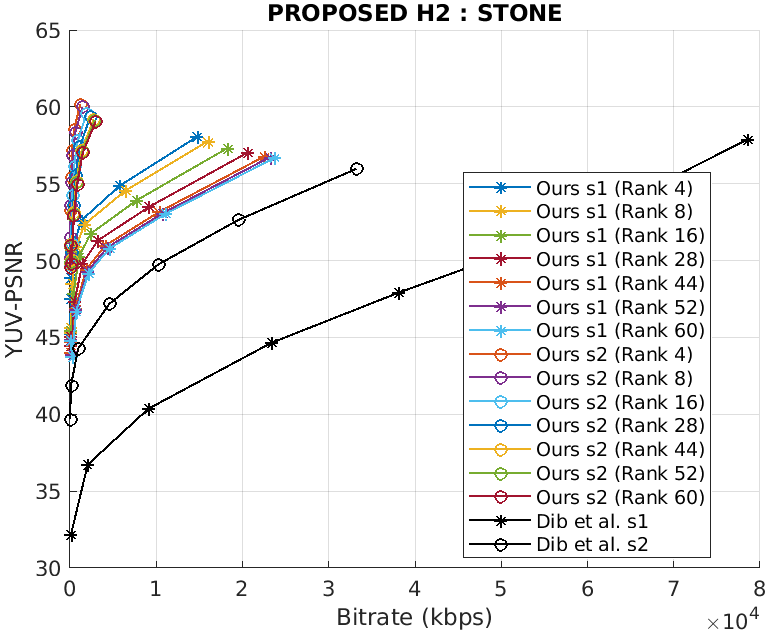}
\end{subfigure} % <-- added

\begin{subfigure}{0.21\textwidth}
  \includegraphics[width=\linewidth]{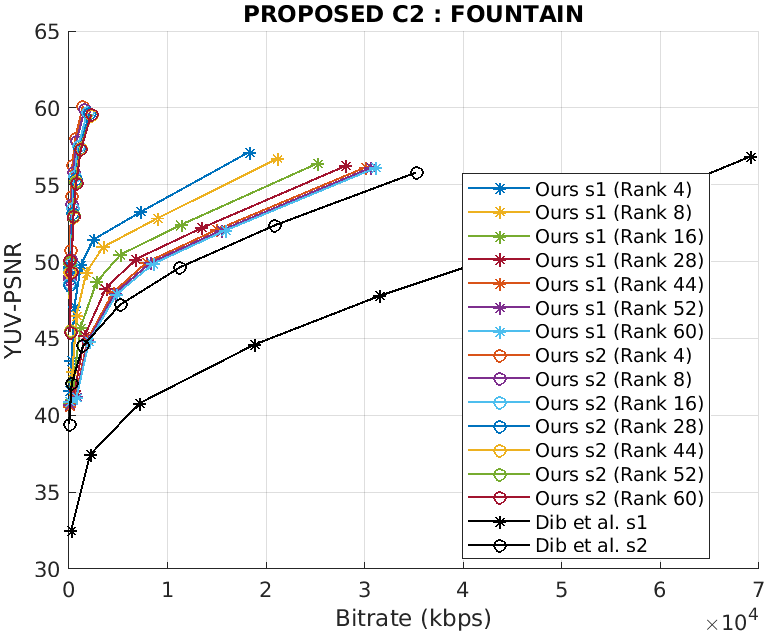}
  %\caption{Reconstructed central SAI,Rank 20, QP 2}
  \caption{\footnotesize Circular-2 }
\end{subfigure} % <-- added
\begin{subfigure}{0.21\textwidth}
  \includegraphics[width=\linewidth]{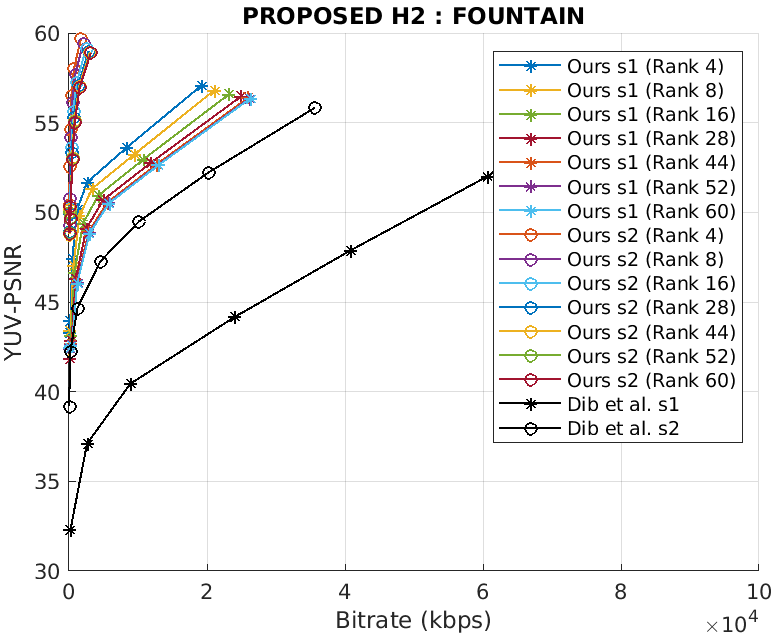}
  %\caption{Reconstructed central SAI,Rank 60, QP 2}
  \caption{\footnotesize Hierarchical-2  }
\end{subfigure}
\caption{\footnotesize Bitrate vs YUV-PSNR plots.}
\label{fig:graphs}
\end{figure}

We compared the proposed scheme with the direct compression of light fields using FDL by Dib et al.~\cite{dib2019light}. The proposed coding scheme outperforms the anchor by large margins. The total number of bytes taken by our approach is comparatively far lesser than the work by Dib et al.~\cite{dib2019light} for all ranks and QPs. These corresponding results are highlighted in Table~\ref{table_bytesC2} for the ($C_2$) pattern and in Table~\ref{table_bytesH2} for ($H_2$) pattern. The bitrate vs YUV-PSNR graphs of the three datasets are illustrated in Fig.~\ref{fig:graphs}. For both the Circular-2 and Hierarchical-2 patterns, the proposed scheme has significantly better rate-distortion results. 

Further, we analysed bitrate reduction (BD-rate) of the proposed scheme with respect to Dib et al.~\cite{dib2019light} using the Bjontgaard metric~\cite{bjontegaard2001calculation}. The average percent difference in rate change was estimated over a range of QPs for the seven chosen ranks. A comparison of the percentage of bitrate savings of our proposed coding scheme with respect to the anchor method for the three chosen light field datasets is shown in Table~\ref{table_bdpsnr}. For the ($C_2$) pattern, the proposed scheme achieves $94.53\%$, $96.39\%$, and $93.96\%$ bitrate reduction compared to Dib et al.~\cite{dib2019light} for the light fields \textit{Bikes}, \textit{Stone pillars outside}, and \textit{Fountain \& Vincent 2} respectively. For the pattern $H_2$, the proposed scheme achieves $97.23\%$, $97.54\%$, and $96.67\%$ bitrate reduction compared to Dib et al. \cite{dib2019light} for the light fields \textit{Bikes}, \textit{Stone pillars outside}, and \textit{Fountain \& Vincent 2} respectively.

%graphs

\section{CONCLUSION}
We have proposed a novel hierarchical coding scheme for light fields based on transmittance patterns of low-rank multiplicative layers and Fourier disparity layers. Typical pseudo sequence-based light field compression schemes~\cite{RwPsvRef1_liu2016pseudo,RwPsvRef3_ahmad2017interpreting,RwPsvRef4_ahmad2019computationally,RwPsvRef5_gu2019high} do not efficiently consider the similarities between horizontal and vertical views of a light field. Our proposed scheme not only exploits the spatial correlation in multiplicative layers of the subset views for varying low ranks, but also removes the temporal inter and intra layer redundancies in the low-rank approximated representation of the view subsets. The approximated light field is further compressed by eliminating intrinsic similarities among neighboring views of the Circular-2 or Hierarchical-2 patterns using Fourier Disparity Layers representation. This integrated compression achieves excellent bitrate reductions without compromising the quality of the reconstructed light field.

Our scheme offers flexibility to cover a range of multiple bitrates using just two trained CNNs to obtain a layered representation of the light field subsets. This critical feature sets our proposed scheme apart from other existing light field coding methods, which usually train a system (or network) to support only specific bitrates during the compression. Besides, existing coding approaches are not explicitly designed to target layered displays and are usually only classified to work for lenslet-based formats or sub-aperture images based pseudo-sequence light field representation. Our coding model can complement existing layered light field displays and is also adaptable to other mobile, head-mounted~\cite{sharma2020unified}, table-top~\cite{maruyama2018implementation} or autostereoscopic displays.

We plan to extend the proposed idea to light field displays with more than three light attenuating layers in our future work. Proof-of-concept experiments with our scheme also pave the way to form a more profound understanding in the rank-analysis of a light-field using other mathematically valid tensor-based models~\cite{wetzstein2012tensor,kobayashi2017focal} and coded mask cameras~\cite{maruyama2019coded}. We also aim to verify our scheme with physical light field display hardware. Further, we wish to adapt the proposed scheme with display device availability and optimize the bandwidth for a target bitrate. This would enable deploying the concepts of layered displays on different display platforms that deliver 3D contents with the limited hardware resources, thus, best meeting the viewers’ preferences for depth impression or visual comfort.

%table3-bdpsnr
\begin{table}[t]
\centering
\caption{\footnotesize Bjontegaard percentage rate savings for the proposed compression scheme wrt Dib et al. \cite{dib2019light}. Negative values represent gains.}
\label{table_bdpsnr}
\resizebox{6cm}{!}{
\makebox[\linewidth]{\begin{tabular}{|c|c|c|c|c|c|}
\hline
 &  & \multicolumn{2}{|c|}{Circular-2} & \multicolumn{2}{|c|}{Hierarchical-2}  \\
\hline
Scene & Rank & Subset 1 & Subset 2 & Subset 1 & Subset 2\\
\hline
\multirow{7}{*}{Bikes} &	4	&	-97.56541214	&	-99.29882593	&	-98.3907291	&	-99.54258115	\\	\cline{2-6}
&	8	&	-95.76892379	&	-99.03123267	&	-97.66246664	&	-99.29956835	\\	\cline{2-6}
&	16	&	-93.17474979	&	-98.45339444	&	-96.63095962	&	-98.95497164	\\	\cline{2-6}
&	28	&	-90.3477664	&	-98.11361299	&	-95.58497187	&	-98.67266268	\\	\cline{2-6}
&	44	&	-87.34955344	&	-97.75859383	&	-94.60404741	&	-98.07437089	\\	\cline{2-6}
&	52	&	-86.18089698	&	-97.59830221	&	-94.16664723	&	-97.93405439	\\	\cline{2-6}
&	60	&	-85.37903466	&	-97.4382535	&	-93.89176465	&	-97.88383496	\\	\cline{1-6}
\hline
\multirow{7}{*}{Stone pillars outside } &	4	&	-98.28738998	&	-99.5671618	&	-98.54643236	&	-99.69494577	\\	\cline{2-6}
&	8	&	-97.48146218	&	-99.45696678	&	-97.88804729	&	-99.65675017	\\	\cline{2-6}
&	16	&	-95.3611157	&	-98.79333481	&	-96.93436888	&	-99.10493755	\\	\cline{2-6}
&	28	&	-93.25557142	&	-98.75966092	&	-95.85758052	&	-98.75257079	\\	\cline{2-6}
&	44	&	-91.33970709	&	-98.53538189	&	-94.86352824	&	-98.5749095	\\	\cline{2-6}
&	52	&	-91.07697181	&	-98.45815322	&	-94.57758068	&	-98.46705075	\\	\cline{2-6}
&	60	&	-90.76161372	&	-98.35636369	&	-94.30087553	&	-98.40819422	\\	\cline{1-6}
\hline
\multirow{7}{*}{Fountain \& Vincent 2} &	4	&	-96.27656828	&	-99.26603902	&	-96.9530291	&	-99.57099646	\\	\cline{2-6}
&	8	&	-94.48002919	&	-98.96179277	&	-96.34841383	&	-99.08720888	\\	\cline{2-6}
&	16	&	-91.14745649	&	-99.02010069	&	-95.17447177	&	-99.23270338	\\	\cline{2-6}
&	28	&	-87.71507969	&	-98.92810522	&	-93.90335982	&	-98.97503776	\\	\cline{2-6}
&	44	&	-85.20018338	&	-98.74257926	&	-92.71888896	&	-98.94441882	\\	\cline{2-6}
&	52	&	-84.44629524	&	-98.77812886	&	-92.50450918	&	-98.83289017	\\	\cline{2-6}
&	60	&	-83.73495541	&	-98.74446465	&	-92.34241435	&	-98.90331735	\\	\cline{1-6}

\hline

\end{tabular} }
}
\end{table}

\addtolength{\textheight}{-12cm}  
\section*{ACKNOWLEDGMENT}
The scientific efforts leading to the results reported in this paper were supported in part by the Department of Science and Technology, Government of India 
under Grant DST/INSPIRE/04/2017/001853.

\bibliographystyle{ieeetr}
\footnotesize
\bibliography{root.bib}

\begin{thebibliography}{10}

\bibitem{li2020light}
T.~Li, Q.~Huang, S.~Alfaro, A.~Supikov, J.~Ratcliff, G.~Grover, and R.~Azuma,
  ``Light-field displays: a view-dependent approach,'' in {\em ACM SIGGRAPH
  2020 ET}, pp.~1--2, 2020.

\bibitem{wetzstein2012tensor}
G.~Wetzstein, D.~R. Lanman, M.~W. Hirsch, and R.~Raskar, ``Tensor displays:
  compressive light field synthesis using multilayer displays with directional
  backlighting,'' 2012.

\bibitem{sharma2016novel}
M.~Sharma, S.~Chaudhury, and B.~Lall, ``A novel hybrid kinect-variety-based
  high-quality multiview rendering scheme for glass-free 3d displays,'' {\em
  IEEE TCSVT}, vol.~27, no.~10, pp.~2098--2117, 2016.

\bibitem{RwlensletRef5_liu2019content}
D.~Liu, P.~An, R.~Ma, W.~Zhan, X.~Huang, and A.~A. Yahya, ``Content-based light
  field image compression method with gaussian process regression,'' {\em IEEE
  TM}, vol.~22, no.~4, pp.~846--859, 2019.

\bibitem{RwlensletRef4_monteiro2017light}
R.~J. Monteiro, P.~J. Nunes, N.~M. Rodrigues, and S.~M. Faria, ``Light field
  image coding using high-order intrablock prediction,'' {\em IEEE JSTSP},
  vol.~11, no.~7, pp.~1120--1131, 2017.

\bibitem{RwlensletRef3_li2016compression}
Y.~Li, R.~Olsson, and M.~Sj{\"o}str{\"o}m, ``Compression of unfocused plenoptic
  images using a displacement intra prediction,'' in {\em 2016 IEEE ICMEW},
  pp.~1--4, IEEE, 2016.

\bibitem{RwDispRef3_dib2020local}
E.~Dib, M.~Le~Pendu, X.~Jiang, and C.~Guillemot, ``Local low rank approximation
  with a parametric disparity model for light field compression,'' {\em IEEE
  TIP}, vol.~29, pp.~9641--9653, 2020.

\bibitem{RwDispRef2_jiang2017light}
X.~Jiang, M.~Le~Pendu, R.~A. Farrugia, and C.~Guillemot, ``Light field
  compression with homography-based low-rank approximation,'' {\em IEEE JSTSP},
  vol.~11, no.~7, pp.~1132--1145, 2017.

\bibitem{RwDispRef1_zhao2017light}
S.~Zhao and Z.~Chen, ``Light field image coding via linear approximation
  prior,'' in {\em IEEE ICIP}, pp.~4562--4566, IEEE, 2017.

\bibitem{RwEpiRef2_ahmad2020shearlet}
W.~Ahmad, S.~Vagharshakyan, M.~Sj{\"o}str{\"o}m, A.~Gotchev, R.~Bregovic, and
  R.~Olsson, ``Shearlet transform-based light field compression under low
  bitrates,'' {\em IEEE TIP}, vol.~29, pp.~4269--4280, 2020.

\bibitem{RwEpiRef3_chen2020light}
Y.~Chen, P.~An, X.~Huang, C.~Yang, D.~Liu, and Q.~Wu, ``Light field compression
  using global multiplane representation and two-step prediction,'' {\em IEEE
  SPL}, vol.~27, pp.~1135--1139, 2020.

\bibitem{RwCbRef1_hu2020adaptive}
X.~Hu, J.~Shan, Y.~Liu, L.~Zhang, and S.~Shirmohammadi, ``An adaptive two-layer
  light field compression scheme using gnn-based reconstruction,'' {\em ACM
  TOMM}, vol.~16, no.~2s, pp.~1--23, 2020.

\bibitem{RwVsRef3_huang2019light}
X.~Huang, P.~An, F.~Cao, D.~Liu, and Q.~Wu, ``Light-field compression using a
  pair of steps and depth estimation,'' {\em Optics express}, vol.~27, no.~3,
  pp.~3557--3573, 2019.

\bibitem{RwVsRef4_heriard2019light}
B.~H{\'e}riard-Dubreuil, I.~Viola, and T.~Ebrahimi, ``Light field compression
  using translation-assisted view estimation,'' in {\em 2019 PCS}, pp.~1--5,
  IEEE, 2019.

\bibitem{RwDeepRef6_schiopu2019deep}
I.~Schiopu and A.~Munteanu, ``Deep-learning-based macro-pixel synthesis and
  lossless coding of light field images,'' {\em APSIPA TSIP}, vol.~8, 2019.

\bibitem{RwDeepRef8_liu2021view}
D.~Liu, X.~Huang, W.~Zhan, L.~Ai, X.~Zheng, and S.~Cheng, ``View
  synthesis-based light field image compression using a generative adversarial
  network,'' {\em Information Sciences}, vol.~545, pp.~118--131, 2021.

\bibitem{RwPsvRef1_liu2016pseudo}
D.~Liu, L.~Wang, L.~Li, Z.~Xiong, F.~Wu, and W.~Zeng, ``Pseudo-sequence-based
  light field image compression,'' in {\em IEEE ICMEW}, pp.~1--4, IEEE, 2016.

\bibitem{RwPsvRef3_ahmad2017interpreting}
W.~Ahmad, R.~Olsson, and M.~Sj{\"o}str{\"o}m, ``Interpreting plenoptic images
  as multi-view sequences for improved compression,'' in {\em IEEE ICIP},
  pp.~4557--4561, IEEE, 2017.

\bibitem{RwPsvRef4_ahmad2019computationally}
W.~Ahmad, M.~Ghafoor, S.~A. Tariq, A.~Hassan, M.~Sj{\"o}str{\"o}m, and
  R.~Olsson, ``Computationally efficient light field image compression using a
  multiview hevc framework,'' {\em IEEE access}, vol.~7, pp.~143002--143014,
  2019.

\bibitem{RwPsvRef5_gu2019high}
J.~Gu, B.~Guo, and J.~Wen, ``High efficiency light field compression via
  virtual reference and hierarchical mv-hevc,'' in {\em IEEE (ICME)},
  pp.~344--349, IEEE, 2019.

\bibitem{maruyama2020comparison}
K.~Maruyama, K.~Takahashi, and T.~Fujii, ``Comparison of layer operations and
  optimization methods for light field display,'' {\em IEEE Access}, vol.~8,
  pp.~38767--38775, 2020.

\bibitem{le2019fourier}
M.~Le~Pendu, C.~Guillemot, and A.~Smolic, ``A fourier disparity layer
  representation for light fields,'' {\em IEEE TIP}, vol.~28, no.~11,
  pp.~5740--5753, 2019.

\bibitem{musco2015randomized}
C.~Musco and C.~Musco, ``Randomized block krylov methods for stronger and
  faster approximate singular value decomposition,'' {\em arXiv preprint
  arXiv:1504.05477}, 2015.

\bibitem{sullivan2012overview}
G.~J. Sullivan, J.-R. Ohm, W.-J. Han, and T.~Wiegand, ``Overview of the high
  efficiency video coding (hevc) standard,'' {\em IEEE TCSVT}, vol.~22, no.~12,
  pp.~1649--1668, 2012.

\bibitem{dib2019light}
E.~Dib, M.~Le~Pendu, and C.~Guillemot, ``Light field compression using fourier
  disparity layers,'' in {\em IEEE ICIP}, pp.~3751--3755, IEEE, 2019.

\bibitem{gortler1996lumigraph}
S.~J. Gortler, R.~Grzeszczuk, R.~Szeliski, and M.~F. Cohen, ``The lumigraph,''
  in {\em Proceedings of the 23rd annual conference on Computer graphics and
  interactive techniques}, pp.~43--54, 1996.

\bibitem{levoy1996light}
M.~Levoy and P.~Hanrahan, ``Light field rendering,'' in {\em Proceedings of the
  23rd annual conference on Computer graphics and interactive techniques},
  pp.~31--42, 1996.

\bibitem{gu1996efficient}
M.~Gu and S.~C. Eisenstat, ``Efficient algorithms for computing a strong
  rank-revealing qr factorization,'' {\em SIAM Journal on Scientific
  Computing}, vol.~17, no.~4, pp.~848--869, 1996.

\bibitem{rerabek2016new}
M.~Rerabek and T.~Ebrahimi, ``New light field image dataset,'' in {\em 8th
  International Conference on QoMEX}, 2016.

\bibitem{dansereau2013decoding}
D.~G. Dansereau, O.~Pizarro, and S.~B. Williams, ``Decoding, calibration and
  rectification for lenselet-based plenoptic cameras,'' in {\em Proceedings of
  the IEEE conference on computer vision and pattern recognition},
  pp.~1027--1034, 2013.

\bibitem{bjontegaard2001calculation}
G.~Bjontegaard, ``Calculation of average psnr differences between rd-curves,''
  {\em VCEG-M33}, 2001.

\bibitem{sharma2020unified}
V.~Thumuluri and M.~Sharma, ``A unified deep learning approach for foveated
  rendering \& novel view synthesis from sparse rgb-d light fields,'' in {\em
  IC3D}, Brussels, Belgium, 2020.

\bibitem{maruyama2018implementation}
K.~Maruyama, H.~Kojima, K.~Takahashi, and T.~Fujii, ``Implementation of
  table-top light-field display,'' in {\em IDW}, 2018.

\bibitem{kobayashi2017focal}
Y.~Kobayashi, K.~Takahashi, and T.~Fujii, ``From focal stacks to tensor
  display: A method for light field visualization without multi-view images,''
  in {\em IEEE ICASSP}, pp.~2007--2011, IEEE, 2017.

\bibitem{maruyama2019coded}
K.~Maruyama, Y.~Inagaki, K.~Takahashi, T.~Fujii, and H.~Nagahara, ``A 3-d
  display pipeline from coded-aperture camera to tensor light-field display
  through cnn,'' in {\em IEEE ICIP}, pp.~1064--1068, IEEE, 2019.

\end{thebibliography}

\end{document}